\documentclass[preprint,12pt]{elsarticle}

\usepackage{graphicx}
\usepackage{amssymb}
\usepackage{amsmath}
\usepackage{url}

\usepackage{lineno}

\journal{Journal of Computational Physics}

\begin{document}

\begin{frontmatter}


\title{Inferring solutions of differential equations using noisy multi-fidelity data}



\author[a]{Maziar Raissi}
\author[b]{Paris Perdikaris} 
\author[a]{George Em Karniadakis}

\address[a]{Division of Applied Mathematics, Brown University, Providence, RI, USA}
\address[b]{Department of Mechanical Engineering, Massachusetts Institute of Technology, Cambridge, MA, USA}

\begin{abstract}
For more than two centuries, solutions of differential equations have been obtained either analytically or numerically based on typically well-behaved forcing and boundary conditions for well-posed problems. We are changing this paradigm in a fundamental way by establishing an interface between probabilistic machine learning and differential equations. We develop data-driven algorithms for general linear equations using Gaussian process priors tailored to the corresponding integro-differential operators. The only observables are scarce noisy multi-fidelity data for the forcing and solution that are not required to reside on the domain boundary. The resulting predictive posterior distributions quantify uncertainty and naturally lead to adaptive solution refinement via active learning. This general framework circumvents the tyranny of numerical discretization as well as the consistency and stability issues of time-integration, and is scalable to high-dimensions.
\end{abstract}

\begin{keyword}
Machine learning \sep Integro-differential equations \sep Multi-fidelity modeling \sep Uncertainty quantification


\end{keyword}

\end{frontmatter}

%\linenumbers

\section{Introduction}
\label{S:1}

Nearly two decades ago a visionary treatise by David Mumford anticipated that "stochastic methods will transform pure and applied mathematics in the beginning of the third millennium, as probability and statistics will come to be viewed as the natural tools to use in mathematical as well as scientific modeling" \cite{mumford2000dawning}.  Indeed, in recent years we have been witnessing the emergence of a data-driven era in which probability and statistics have been the focal point in the development of disruptive technologies such as probabilistic machine learning \cite{ghahramani2015probabilistic,jordan2015machine}. Only to verify Mumford's predictions, this wave of change is steadily propagating into applied mathematics, giving rise to novel probabilistic interpretations of classical deterministic scientific methods and algorithms. This new viewpoint offers an elegant path to generalization and enables computing with probability distributions rather than solely relying on deterministic thinking. In particular, in the area of numerical analysis and scientific computing, the first hints of this paradigm shift were clearly manifested in the thought-provoking work of Diaconis \cite{diaconis1988bayesian}, tracing back to Poincaré's courses on probability theory \cite{poincare1912calcul}. This line of work has recently inspired resurgence in probabilistic methods and algorithms \cite{hennig2015probabilistic,owhadi2015bayesian,sarkka2011linear} that offer a principled and robust handling of uncertainty due to model inadequacy, parametric uncertainties, and numerical discretization/truncation errors. These developments are defining a new area of scientific research in which probabilistic machine learning and classical scientific computing coexist in unison, providing a flexible and general platform for Bayesian reasoning and computation. In this work, we exploit this interface by developing Bayesian inference algorithms that are able to learn from data and equations in a synergistic fashion.

\section{Problem setup}

We consider general linear integro-differential equations of the form 
$$
\mathcal{L}_x u(x) = f(x),
$$
where $x$ is a $D$-dimensional vector that includes spatial or temporal coordinates, $\mathcal{L}_x$ is a linear operator, $u(x)$ denotes an unknown solution to the equation, and $f(x)$ represents the external force that drives the system. We assume that $f_L:=f$ is a complex, expensive to evaluate,``€œblack-box"€ function. For instance, $f_L$ could represent force acting upon a physical system, the outcome of a costly experiment, the output of an expensive computer code, or any other unknown function. We assume limited availability of high-fidelity data for $f_L$, denoted by $\{\mathbf{x}_L,\mathbf{y}_L\}$, that could be corrupted by noise $\mathbf{\epsilon}_L$, i.e., $\mathbf{y}_L = f_L(\mathbf{x}_L) + \mathbf{\epsilon}_L$. In many cases, we may also have access to supplementary sets of less accurate models $f_\ell, \ell=1,\ldots,L-1$, sorted by increasing level of fidelity, and generating data $\{\mathbf{x}_\ell,\mathbf{y}_\ell\}$ that could also be contaminated by noise $\mathbf{\epsilon}_\ell$, i.e., $\mathbf{y}_\ell = f_\ell(\mathbf{x}_\ell) + \mathbf{\epsilon}_\ell$. Such data may come from simplified computer models, inexpensive sensors, or uncalibrated measurements. In addition, we also have a small set of data on the solution $u$, denoted by $\{\mathbf{x}_0,\mathbf{y}_0\}$, perturbed by noise $\mathbf{\epsilon}_0$, i.e., $\mathbf{y}_0 = u(\mathbf{x}_0) + \mathbf{\epsilon}_0$, sampled at scattered spatio-temporal locations, which we call anchor points to distinguish them from boundary or initial values. Although they could be located on the domain boundaries as in the classical setting, this is not a requirement in the current framework as solution data could be partially available on the boundary or in the interior of either spatial or temporal domains. Here, we are not primarily interested in estimating $f$. We are interested in estimating the unknown solution $u$ that is related to $f$ through the linear operator $\mathcal{L}_x$.
For example, consider a bridge subject to environmental loading. In a two-level of fidelity setting (i.e., $L=2$), suppose that one could only afford to collect scarce but accurate (high-fidelity) measurements of the wind force $f_2(x)$ acting upon the bridge at some locations. In addition, one could also gather samples by probing a cheaper but inaccurate (low-fidelity) wind prediction model $f_1(x)$ at some other locations. How could this noisy data be combined to accurately estimate the bridge displacements $u(x)$ under the laws of linear elasticity? What is the uncertainty/error associated with this estimation? How can we best improve that estimation if we can afford another observation of the wind force?
Quoting Diaconis \cite{diaconis1988bayesian}, ``once we allow that we don't know $f$, but do know some things, it becomes natural to take a Bayesian approach".

\section{Solution methodology}
\begin{figure}
\centering
\includegraphics[width=\textwidth]{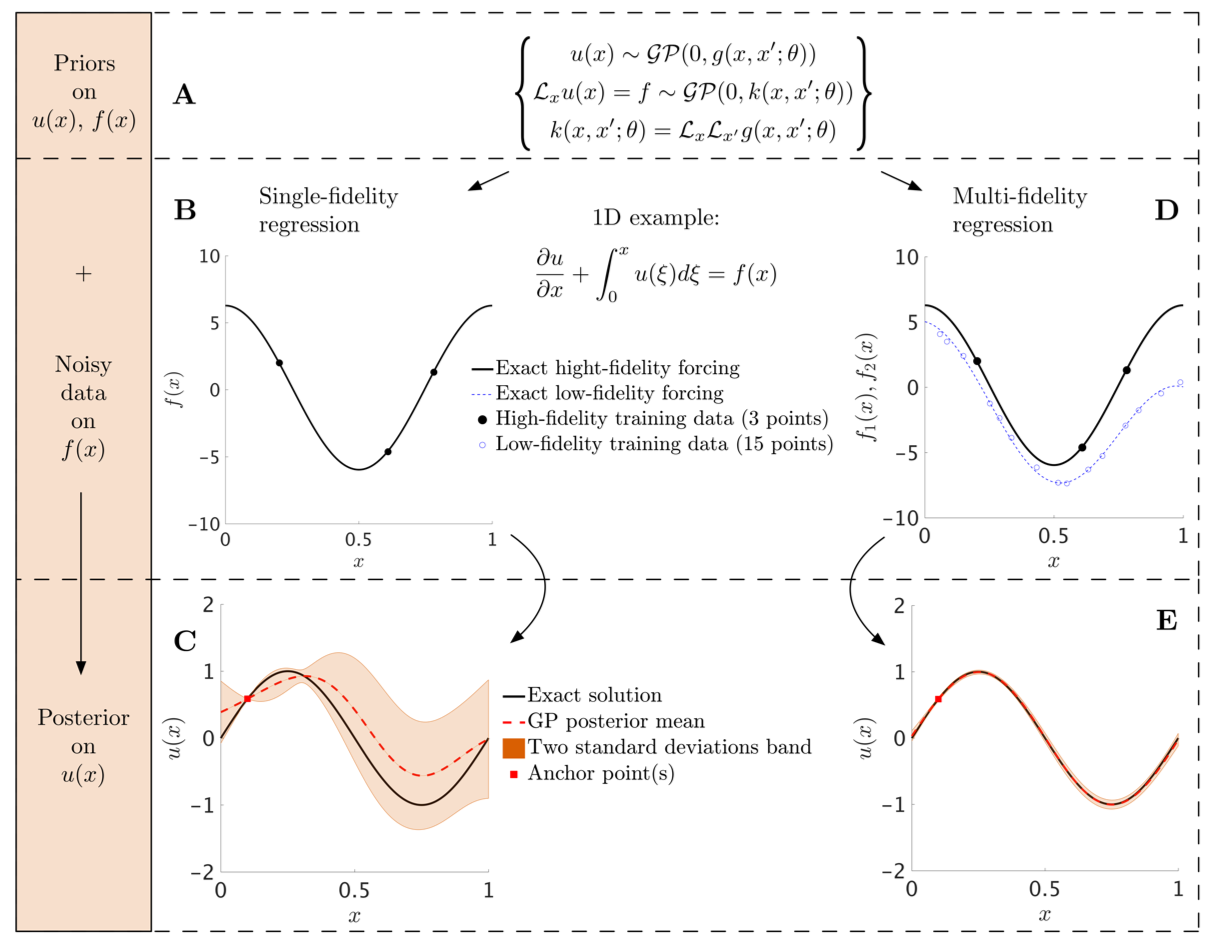}
\caption{{\em Inferring solutions of differential equations using noisy multi-fidelity data}: (A) Starting from a GP prior on $u$ with kernel $g(x,x';\theta)$, and using the linearity of the operator $\mathcal{L}_x$, we obtain a GP prior on $f$ that encodes the structure of the differential equation in its covariance kernel $k(x,x';\theta)$. (B) In view of 3 noisy high-fidelity data points for $f$, we train a single-fidelity GP (i.e., $\rho=0$) with kernel $k(x,x';\theta)$ to estimate the hyper-parameters $\theta$. (C) This leads to a predictive posterior distribution for $u$ conditioned on the available data on $f$ and the anchor point(s) on $u$. For instance, in the one-dimensional integro-differential example considered here, the posterior mean gives us an estimate of the solution $u$ while the posterior variance quantifies uncertainty in our predictions. (D), (E) Adding a supplementary set of 15 noisy low-fidelity data points for $f$, and training a multi-fidelity GP, we obtain significantly more accurate solutions with a tighter uncertainty band.}
\label{fig:fig1}
\end{figure}

The basic building blocks of the Bayesian approach adopted here are Gaussian process (GP) regression \cite{Rasmussen06gaussianprocesses,murphy2012machine} and auto-regressive stochastic schemes \cite{kennedy2000predicting,le2013multi}. This choice is motivated by the Bayesian non-parametric nature of GPs, their analytical tractability properties, and their natural extension to the multi-fidelity settings that are fundamental to this work. In particular, GPs provide a flexible prior distribution over functions, and, more importantly, a fully probabilistic workflow that returns robust posterior variance estimates which enable adaptive refinement and active learning \cite{cohn1996active,krause2007nonmyopic,mackay1992information}. The framework we propose is summarized in Figure \ref{fig:fig1} and is outlined in the following.

Inspired by \cite{kennedy2000predicting,le2013multi}, we will present the framework considering two-levels of fidelity (i.e. $L=2)$, although generalization to multiple levels is straightforward. Let us start with the auto-regressive model $u(x) = \rho u_1(x) + \delta_2(x)$, where $\delta_2(x)$ and $u_1(x)$ are two independent Gaussian processes \cite{Rasmussen06gaussianprocesses,murphy2012machine,kennedy2000predicting,le2013multi} with $
\delta_2(x) \sim \mathcal{GP}(0,g_2(x,x';\theta_2))$ and $u_1(x) \sim \mathcal{GP}(0,g_1(x,x';\theta_1))$. Here, $g_1(x,x';\theta_1), g_2(x,x';\theta_2)$ are covariance functions, $\theta_1, \theta_2$ denote their hyper-parameters, and $\rho$ is a cross-correlation parameter to be learned from the data (see Sec.~\ref{sec:NLML}). Then, one can trivially obtain
$$
u(x) \sim \mathcal{GP}(0,g(x,x';\theta)),
$$
with $g(x,x';\theta) = \rho^2 g_1(x,x';\theta_1) + g_2(x,x';\theta_2)$, and $\theta = (\theta_1,\theta_2,\rho)$. The key observation here is that the derivatives and integrals of a Gaussian process are still Gaussian processes. Therefore, given that the operator $\mathcal{L}_x$ is linear, we obtain
$$
f(x) \sim \mathcal{GP}(0, k(x,x';\theta)),
$$
with
$$
k(x,x';\theta) = \mathcal{L}_x \mathcal{L}_{x'} g(x,x';\theta).
$$
Similarly, we arrive at the auto-regressive structure $f(x) = \rho f_1(x) + \gamma_2(x)$ on the forcing, where $\gamma_2(x) = \mathcal{L}_x \delta_2(x)$, and $f_1(x) = \mathcal{L}_x u_1(x)$ are consequently two independent Gaussian processes with $\gamma_2(x) \sim \mathcal{GP}(0,k_2(x,x';\theta_2))$, $f_1(x) \sim \mathcal{GP}(0,k_1(x,x';\theta_1))$. Furthermore, for $\ell = 1,2$, $k_\ell(x,x';\theta_\ell) = \mathcal{L}_x \mathcal{L}_{x'} g_\ell(x,x';\theta_\ell)$. The hyper-parameters $\theta = (\theta_1,\theta_2,\rho)$ which are shared between the kernels $g(x,x';\theta)$ and $k(x,x';\theta)$ can be estimated by minimizing the negative log marginal likelihood (see Sec.~\ref{sec:NLML})
$$
\mathcal{NLML}(\theta,\sigma_{n_0}^{2},\sigma_{n_1}^{2},\sigma_{n_2}^{2}) := -\log p(\mathbf{y} | \mathbf{x};\theta,\sigma_{n_0}^{2},\sigma_{n_1}^{2},\sigma_{n_2}^{2}),
$$
with $\mathbf{y}^T:=\left[\mathbf{y}_0^T\ \mathbf{y}_1^T\  \mathbf{y}_2^T\right]$ and $\mathbf{x}^T:=\left[\mathbf{x}_0^T\ \mathbf{x}_1^T\  \mathbf{x}_2^T\right]$. Also, the variance parameters associated with the observation noise in 
$u(x),f_1(x)$ and $f_2(x)$ are denoted by $\sigma_{n_0}^{2},\sigma_{n_1}^{2}$, and $\sigma_{n_2}^{2}$, respectively.  Once the model has been trained on the available multi-fidelity data on $f$ and anchor points on $u$, we obtain a GP posterior distribution on $u$ with predictive mean $\overline{u}$ which can be used to perform predictions at a new test point with quantified uncertainty (see Sec.~\ref{sec:prediction}). The most computationally intensive part of this workflow is associated with inverting dense covariance matrices $K$ during model training, and scales cubically with the number of training data (see Sec.~\ref{sec:cost}).

\subsection{Training}\label{sec:NLML}
The hyper-parameters $\theta = (\theta_1,\theta_2,\rho)$ which are shared between the kernels $g(x,x';\theta)$ and $k(x,x';\theta)$ can be estimated by minimizing the negative log marginal likelihood
\[
\mathcal{NLML}(\theta,\sigma_{n_0}^{2},\sigma_{n_1}^{2},\sigma_{n_2}^{2}) := -\log p(\mathbf{y} | \mathbf{x};\theta,\sigma_{n_0}^{2},\sigma_{n_1}^{2},\sigma_{n_2}^{2}),
\]
with $\mathbf{y}^T:=\left[\mathbf{y}_0^T\ \mathbf{y}_1^T\  \mathbf{y}_2^T\right]$ and $\mathbf{x}^T:=\left[\mathbf{x}_0^T\ \mathbf{x}_1^T\  \mathbf{x}_2^T\right]$. Also, the variance parameters associated with the observation noise in 
$u(x),f_1(x)$ and $f_2(x)$ are denoted by $\sigma_{n_0}^{2},\sigma_{n_1}^{2}$, and $\sigma_{n_2}^{2}$, respectively. Finally, the negative log marginal likelihood is explicitly given by
\[
\mathcal{NLML} = \frac{1}{2} \mathbf{y}^T K^{-1} \mathbf{y} + \frac{1}{2} \log |K| +\frac{n}{2}\log (2\pi),
\]
where $n = n_0 + n_1 + n_2$, denotes the total number of data points in $\mathbf{x}^T:=\left[\mathbf{x}_0^T\ \mathbf{x}_1^T\  \mathbf{x}_2^T\right]$, and
\[
K = \left[ \begin{array}{ccc}
K_{00} & K_{01} & K_{02} \\
K_{10} & K_{11} & K_{12} \\
K_{20} & K_{21} & K_{22}
\end{array} \right],
\]
and 
\begin{align*}
K_{00} & = g(\mathbf{x}_{0},\mathbf{x}_{0}; \theta)+ \sigma_{n_0}^2 I_0,  \\ 
K_{01} & = K_{10}^T = \rho \mathcal{L}_{x'} g_1(\mathbf{x}_0,\mathbf{x}_1; \theta_1), \\ 
K_{02} & = K_{20}^T =\mathcal{L}_{x'} g(\mathbf{x}_0,\mathbf{x}_2; \theta_2), \\ 
K_{11} & = k_1(\mathbf{x}_1,\mathbf{x}_1;\theta_1) + \sigma_{n_1}^2 I_1, \\ 
K_{12} & = K_{21}^T = \rho k_1(\mathbf{x}_1,\mathbf{x}_2;\theta_1), \\ 
K_{22} & = k(\mathbf{x}_2,\mathbf{x}_2;\theta) + \sigma_{n_2}^2 I_2,
\end{align*}
with $I_0, I_1$, and $I_2$ being the identity matrices of size $n_0, n_1$, and $n_2$, respectively.

\subsection{Kernels}\label{sec:kernels}
Without loss of generality, all Gaussian process priors used in this work are assumed to have zero mean and a squared exponential covariance function \cite{Rasmussen06gaussianprocesses,murphy2012machine,kennedy2000predicting,le2013multi}. 
Moreover, anisotropy across input dimensions is handled by Automatic Relevance Determination (ARD) weights $w_{d,\ell}$ \cite{Rasmussen06gaussianprocesses}
\[
g_\ell(x,x';\theta_\ell) = \sigma_{\ell}^2 \exp\left(-\frac12\sum_{d=1}^D w_{d,\ell}(x_d - x'_d)^2\right), \ \ \text{for} \ \ \ell=1,2,
\]	
where $\sigma_\ell^2$ is a variance parameter, $x$ is a $D$-dimensional vector that includes spatial or temporal coordinates, and $\theta_\ell = \left(\sigma_{\ell}^{2},\left(w_{d,\ell}\right)_{d=1}^D\right)$. The choice of the kernel represents our prior belief about the properties of the functions we are trying to approximate. From a theoretical point of view, each kernel gives rise to a Reproducing Kernel Hilbert Space \cite{Rasmussen06gaussianprocesses} 
that defines the class of functions that can be represented by our prior. In particular, the squared exponential covariance function chosen above, implies that we are seeking smooth approximations. More complex function classes can be accommodated by appropriately choosing kernels.

\subsection{Cross-correlation parameter}\label{sec:rho}
If the training procedure yields a $\rho$ close to zero, this indicates a negligible cross-correlation between the low- and high-fidelity data. Essentially, this implies that the low-fidelity data is not informative, and the algorithm will automatically ignore them, thus solely trusting the high-fidelity data. In general, $\rho$ could depend on $x$ (i.e., $\rho(x)$), yielding a more expressive scheme that can capture increasingly complex cross correlations \cite{le2013multi}. 
However, for the sake of clarity, this is not pursued in this work.

\subsection{Prediction}\label{sec:prediction}
After training the model on data $\{\mathbf{x}_2,\mathbf{y}_2\}$ on $f_2$, $\{\mathbf{x}_1,\mathbf{y}_1\}$ on $f_1$, and anchor points data $\{\mathbf{x}_0,\mathbf{y}_0\}$ on $u$, we are interested in predicting the value $u(x)$ at a new test point $x$. Hence, we are interested in the posterior distribution $p(u(x) | \mathbf{y}).$ This can be computed by first observing that
\[
\left[\begin{array}{c}
u(x) \\ 
\mathbf{y}
\end{array} \right] \sim \mathcal{N}\left( \left[\begin{array}{c}
0 \\ 
\mathbf{0}
\end{array} \right], \left[ \begin{array}{cc}
g(x,x) & \mathbf{a} \\ 
\mathbf{a}^T & K
\end{array}   \right] \right),
\]
where $\mathbf{a} := \left[ g(x,\mathbf{x}_0; \theta) ~,~ \rho \mathcal{L}_{x'} g_1(x^*,\mathbf{x}_1; \theta_1) ~,~ \mathcal{L}_{x'} g(x^*,\mathbf{x}_2; \theta) \right].$ Therefore, we obtain the predictive distribution $p(u(x) | \mathbf{y}) = \mathcal{N}\left( \overline{u}(x), \ \ V_u(x) \right),$ with predictive mean $\overline{u}(x) := \mathbf{a} K^{-1} \mathbf{y}$ and predictive variance $V_u(x) := g(x,x) - \mathbf{a} K^{-1}\mathbf{a}^T$.

\subsection{Computational cost}\label{sec:cost}
The training step scales as $\mathcal{O}(m n^3)$, where $m$ is the number of optimization iterations needed. In our implementation, we have derived the gradients of the likelihood with respect to all unknown parameters and hyper-parameters \cite{Rasmussen06gaussianprocesses}), 
and used a Quasi-Newton optimizer L-BFGS \cite{liu1989limited} 
with randomized initial guesses. Although this scaling is a well-known limitation of Gaussian process regression, we must emphasize that it has been effectively addressed by the recent works of Snelson \& Gharhamani, and Hensman \& Lawrence \cite{snelson2005sparse,hensman2013gaussian}, 
and by the recursive multi-fidelity scheme put forth by Le Gratiet and Garnier \cite{le2013multi}. 
Finally, we employ $\overline{u}(x)$ to predict $u(x)$ at a new test point $x$ with a linear cost.

\subsection{Adaptive refinement via active learning}
Here we provide details on adaptive acquisition of data in order to enhance our knowledge about the solution $u$, under the assumption that we can afford one additional high-fidelity observation of the right-hand-side forcing $f$. The adaptation of the following active learning scheme to cases where one can acquire additional anchor points or low-fidelity data for $f_1$ is straightforward. We start by obtaining the following predictive distribution for $f(x)$ at a new test point $x$,
$p(f(x) | \mathbf{y}) = \mathcal{N}(\overline{f}(x),V_f(x)),$ where $\overline{f}(x) = \mathbf{b} K^{-1} \mathbf{y}$, $V_f(x) = k(x,x) - \mathbf{b} K^{-1} \mathbf{b}^T$, and $\mathbf{b} := \left[\mathcal{L}_x g(x,\mathbf{x}_0;\theta), \rho \mathcal{L}_x g_1(x,\mathbf{x}_1;\theta_1), k(x,\mathbf{x}_2;\theta) \right].$ The most intuitive sampling strategy corresponds to adding a new observation $x^*$ for $f$ at the location where the posterior variance is maximized, i.e.,
\[
x^* = \arg \max_x V_f(x).
\]
Compared to more sophisticated data acquisition criteria \cite{cohn1996active,krause2007nonmyopic,mackay1992information},
we found that this simple computationally inexpensive choice leads to similar performance for all cases examined in this work. Designing the optimal data acquisition policy for a given problem is still an open question \cite{cohn1996active,krause2007nonmyopic,mackay1992information}.

\section{Results}

\subsection{Integro-differential equation in 1D}
We start with a pedagogical example involving the following one dimensional integro-differential equation 

$$
\frac{\partial}{\partial x}u(x) + \int_0^x u(\xi) d\xi = f(x),
$$
and assume that the low- and high-fidelity training data $\{\mathbf{x}_{1},\mathbf{y}_{1}\},\{\mathbf{x}_2,\mathbf{y}_2\}$ are generated  according to $\mathbf{y}_\ell = f_\ell(\mathbf{x}_\ell) + \mathbf{\epsilon_i}$, $\ell=1,2$,
where $\mathbf{\epsilon_1} \sim \mathcal{N}(0,0.3 I)$,  $\mathbf{\epsilon_2} \sim \mathcal{N}(0,0.05 I)$, $f_2(x) = 2\pi\cos(2\pi{x})+\frac{1}{\pi}\sin(\pi{x})^{2}$, and $f_1(x) = 0.8 f_2(x) - 5 x$. This induces a non-trivial 
 cross-correlation structure between $f_1(x),f_2(x)$. The training data points $\mathbf{x}_1$ and $\mathbf{x}_2$ are randomly chosen in the interval $[0,1]$ according to a uniform distribution. Here we take $n_1 = 15$ and $n_2 = 3$, where $n_1$ and $n_2$ denote the sample sizes of $\mathbf{x}_1$ and $\mathbf{x}_2$, respectively. Moreover, we have access to anchor point data $\{x_0,u_0\}$ on $u(x)$. For this example, we randomly selected $x_0 = 0$ in the interval $[0,1]$ and let $y_0 = u(x_0)$. Notice that $u(x) = \sin (2\pi x)$ is the exact solution to the problem. Figure 1 of the manuscript summarizes the results corresponding to: 1) Single-fidelity data for $f$, i.e., $n_1=0$ and $n_2=3$, and 2) Multi-fidelity data for $f_1$ and $f_2$, i.e., $n_1=15$ and $n_2=3$, respectively.

Figure \ref{fig:fig1} highlights the ability of the proposed methodology to accurately approximate the solution to a one dimensional integro-differential equation (see Figure  \ref{fig:fig1}) in the absence of any numerical discretization of the linear operator, or any data on $u$ other than the minimal set of anchor points that are necessary to pin down a solution. In sharp contrast to classical grid-based solution strategies (e.g. finite difference, finite element methods, etc.), our machine learning approach relaxes the classical well-possedness requirements as the anchor point(s) need not necessarily be prescribed as initial/boundary conditions, and could also be contaminated by noise. Moreover, we see in Figure \ref{fig:fig1}(C), (E) that a direct consequence of our Bayesian approach is the built-in uncertainty quantification encoded in the posterior variance of $u$. The computed variance reveals regions where model predictions are least trusted, thus directly quantifying model inadequacy. This information is very useful in designing a data acquisition plan that can be used to optimally enhance our knowledge about $u$. Giving rise to an iterative procedure often referred to as active learning \cite{cohn1996active,krause2007nonmyopic,mackay1992information}, this observation can be used to efficiently learn solutions to differential equations by intelligently selecting the location of the next sampling point.

\subsection{Active learning and a-posteriori error estimates for the 2D Poisson equation}

\begin{figure}
\centering
\includegraphics[width=\textwidth]{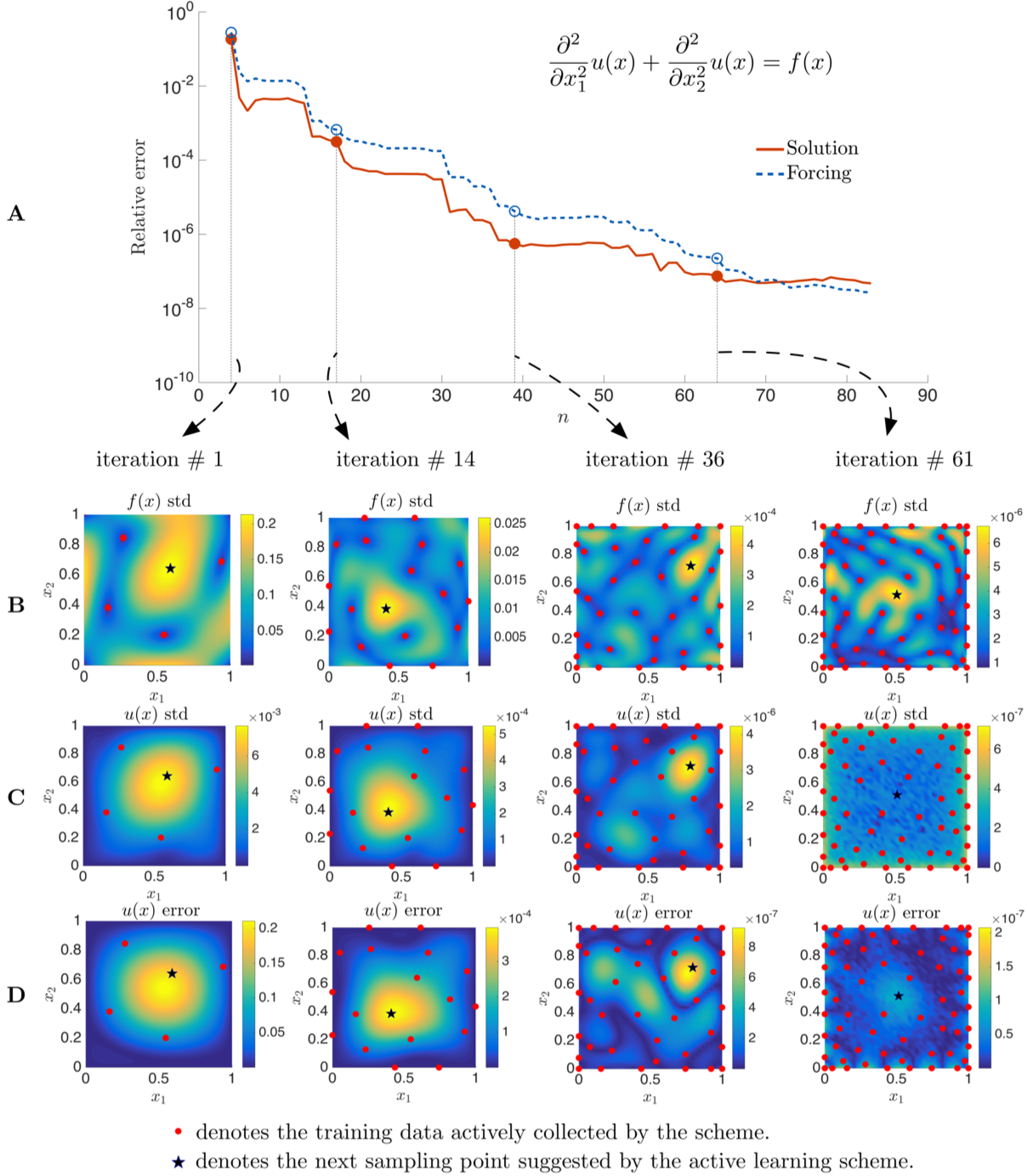}
\caption{{\em Active learning of solutions to linear differential equations and a-posteriori error estimates}: (A) Log-scale convergence plot of the relative error in the predicted solution $u$ and forcing term $f$ as the number of single-fidelity training data on $f$ is increased via active learning. Our point selection policy is guided by the maximum posterior uncertainty on $f$. (B) Evolution of the posterior standard deviation of $f$ as the number of active learning iterations is increased. (C), (D) Evolution of the posterior standard deviation of $u$ and the relative point-wise error against the exact solution. A visual inspection demonstrates the ability of the proposed methodology to provide an {\em a-posteriori error estimate} on the predicted solution. Movie S1 presents a real-time animation of the active learning loop and corresponding convergence.}
\label{fig:fig2}
\end{figure}

Consider the following differential equation

$$
\dfrac{\partial^2}{\partial{x_1}^2}u(x) + \dfrac{\partial^2}{\partial{x_2}^2}u(x) = f(x),
$$
and a single-fidelity data-set comprising of noise-free observations for the forcing 
term $f(x) = -2\pi^2 \sin(\pi x_1) \sin(\pi x_2)$, along with noise free anchor points generated by the exact solution $u(x) = \sin(\pi x_1)\sin(\pi x_2)$. To demonstrate the concept of active learning we start from an initial data set consisting of $4$ randomly sampled observations of $f$ in the unit square, along with 25 anchor points per domain boundary. The latter can be considered as information that is a-priori known from the problem setup, as for this problem we have considered a noise-free Dirichlet boundary condition. Moreover, this relatively high-number of anchor points allows us to accurately resolve the solution on the domain boundary and focus our attention on the convergence properties of our scheme in the interior domain. Starting from this initial training set, we enter an active learning iteration loop in which a new observation of $f$ is augmented to our training set at each iteration according to the chosen sampling policy. Here, we have chosen the most intuitive sampling criterion, namely adding new observations at the locations for which the posterior variance of $f$ is the highest. Despite its simplicity, this choice yields a fast convergence rate, and returns an accurate prediction for the solution $u$ after just a handful of iterations (see Figure \ref{fig:fig2}(A)). Interestingly, the error in the solution seems to be bounded by the approximation error in the forcing term, except for the late iteration stages where the error is dominated by how well we approximate the solution on the boundary. This indicates that in order to further reduce the relative error in $u$, more anchor points on the boundary are needed. Overall, the non-monotonic error decrease observed in Figure \ref{fig:fig2}(A) is a common feature of active learning approaches as the algorithm needs to explore and gather the sufficient information required in order to further reduce the error. Lastly, note how uncertainty in computation is quantified by the computed posterior variance that can be interpreted as a type of {\em a-posteriori} error estimate (see Figure \ref{fig:fig2}(C, D)).

\subsection{Generality and scalability of the method}
It is important to emphasize that as long as the equations are linear, the observations made so far are not problem specific. In fact, the proposed algorithm provides an entirely {\em agnostic} treatment of linear operators, which can be of fundamentally different nature. For example, we can seamlessly learn solutions to integro-differential, time-dependent, high-dimensional, or even fractional equations. This generality and scalability is illustrated through a mosaic of benchmark problems compiled in Figure \ref{fig:fig3}.

\subsubsection{Time-dependent linear advection-diffusion-reaction equation} 
This example is chosen to highlight the capability of the proposed framework to handle time-dependent problems using only noisy scattered space-time observations of the right-hand-side forcing term. To illustrate this capability we consider a time-dependent advection-diffusion-reaction equation 

$$
\frac{\partial}{\partial t} u(t,x) + \frac{\partial}{\partial x} u(t,x) - \frac{\partial^2}{\partial x^2} u(t,x) - u(t,x) = f(x).
$$
Here, we generate a total of $n_1 = 30$ low-fidelity and $n_2 = 10$ high-fidelity training points $(\mathbf{t}_1,\mathbf{x}_1)$ and $(\mathbf{t}_2,\mathbf{x}_2)$, respectively, in the domain $[0,1]^2 = \left\{(t,x): t\in[0,1] \text{ and } x\in[0,1]\right\}$. These points are chosen at random according to a uniform distribution. The low- and high-fidelity training data $\{(\mathbf{t}_1,\mathbf{x}_1), \mathbf{y}_1\},\{(\mathbf{t}_2,\mathbf{x}_2),\mathbf{y}_2\}$ are given by $\mathbf{y}_\ell = f_\ell(\mathbf{t}_{\ell},\mathbf{x}_\ell) + \mathbf{\epsilon}_\ell$, $\ell=1,2,$ where $f_2(t,x) = e^{-t} \left(2 \pi \cos(2 \pi x) + 2 (2 \pi^2 - 1) \sin(2 \pi) x)\right)$, and $f_1(t,x) = 0.8 f_2(t,x) - 5 t x - 20$. Moreover, $\mathbf{\epsilon}_1 \sim \mathcal{N}(0,0.3\ I)$ and $\mathbf{\epsilon}_2 \sim \mathcal{N}(0,0.05\ I)$. We choose $n_0 = 10$ random anchor points $(\mathbf{t}_0,\mathbf{x}_0)$ according to a uniform distribution on the initial/boundary set $\{0\}\times [0,1] \cup [0,1]\times \{0,1\}$. Moreover, $\mathbf{y}_0 = u(\mathbf{t}_0,\mathbf{x}_0) + \mathbf{\epsilon}_0$ with $\mathbf{\epsilon}_0 \sim \mathcal{N}(0,0.01\ I)$. Note that $u(t,x) = e^{-t} \sin(\pi x)$ is the exact solution.

Remarkably, the proposed method circumvents the need for temporal discretization, and is essentially immune to any restrictions arising due to time-stepping, e.g., the fundamental consistency and stability issues in classical numerical analysis. As shown in Figure \ref{fig:fig3}(A), a reasonable reconstruction of the solution field $u$ can be achieved using only 10 noisy high-fidelity observations of the forcing term $f$ (see Figure \ref{fig:fig3}(A-1, A-2)). More importantly, the maximum error in the prediction is quantified by the posterior variance (see Figure \ref{fig:fig3}(A-3)), which, in turn, is in good agreement with the maximum absolute point-wise error between the predicted and exact solution for $u$ (see Figure \ref{fig:fig3}(A-4)). Note that in realistic scenarios no knowledge of the exact solution is available, and therefore one cannot assess model accuracy or inadequacy. The merits of our Bayesian approach are evident -- using a very limited number of noisy high-fidelity observations of $f$ we are able to compute a reasonably accurate solution $u$ avoiding any numerical discretization of the spatio-temporal advection-diffusion-reaction operator.

\subsubsection{Poisson equation in 10D}
To demonstrate scalability to high dimensions, next we consider a 10-dimensional (10D) Poisson equation (see Figure \ref{fig:fig3}(B)) for which only two dimensions are active in the variability of its solution, namely dimensions 1 and 3.  To this end, consider the following differential equation

$$
\sum_{d=1}^{10} \frac{\partial^{2}}{\partial x_d^{2}}u(x) = f(x).
$$
We assume that the low- and high-fidelity data $\{\mathbf{x}_1,\mathbf{y}_1\}$, $\{\mathbf{x}_2,\mathbf{y}_2\}$ are generated according to $\mathbf{y}_\ell = f_\ell(\mathbf{x}_\ell) + \mathbf{\epsilon}_\ell$, $\ell=1,2$, where $\mathbf{\epsilon}_1 \sim \mathcal{N}(0, 0.3\ I)$ and $\mathbf{\epsilon}_2 \sim \mathcal{N}(0, 0.05\ I)$. We construct a training set consisting of $n_1=60$ low-fidelity and $n_2=20$ high-fidelity observations, sampled at random in the unit hyper-cube $[0,1]^{10}$. Moreover, we employ $n_0=40$ data points on the solution $u(x)$. These anchor points are not necessarily boundary points and are in fact randomly chosen in the domain $[0,1]^{10}$ according to a uniform distribution. The high- and low-fidelity forcing terms are given by $f_2(x)=-8 \pi^2 \sin(2\pi x_1)\sin(2\pi x_3)$, and $f_1(x) = 0.8 f_2(x) - 40 \prod_{d=1}^{10} x_d + 30$, respectively. Once again, the data $\mathbf{y}_0$ on the exact solution $u(x) = \sin(2\pi x_1) \sin(2\pi x_3)$ are generated by $\mathbf{y}_0 = u(\mathbf{x}_0) + \mathbf{\epsilon}_0$ with $\mathbf{\epsilon}_0 \sim \mathcal{N}(0,0.01\ I)$. It should be emphasized that the effective dimensionality of this problem is 2, and the active dimensions $x_1$ and $x_3$ will be automatically discovered by our method.

Our goal here is to highlight an important feature of the proposed methodology, namely automatic discovery of this {\em effective dimensionality} from data. This screening procedure is implicitly carried out during model training by using GP covariance kernels that can detect directional anisotropy in multi-fidelity data, helping the algorithm to automatically detect and exploit any low-dimensional structure.  Although the high-fidelity forcing $f_2$ only contains terms involving dimensions 1 and 3, the low-fidelity model $f_1$ is active along all dimensions.  Figure \ref{fig:fig3}(B-1, B-2, B-3) provides a visual assessment of the high accuracy attained by the predictive mean in approximating the exact solution $u$ evaluated at randomly chosen validation points in the 10-dimensional space. Specifically, Figure \ref{fig:fig3}(B-1) is a scatter plot of the predictive mean, Figure \ref{fig:fig3}(B-2) depicts the histogram of the predicted solution values versus the exact solution, and Figure \ref{fig:fig3}(B-3) is a one dimensional slice of the solution field. If all the dimensions are active, achieving this accuracy level would clearly require a larger number of multi-fidelity training data. However, the important point here is that our algorithm can discover the effective dimensionality of the system from data (see Figure \ref{fig:fig3}(B-4)), which is a non-trivial problem.

\subsubsection{Fractional sub-diffusion equation}

\begin{figure}
\centering
\includegraphics[width=\textwidth]{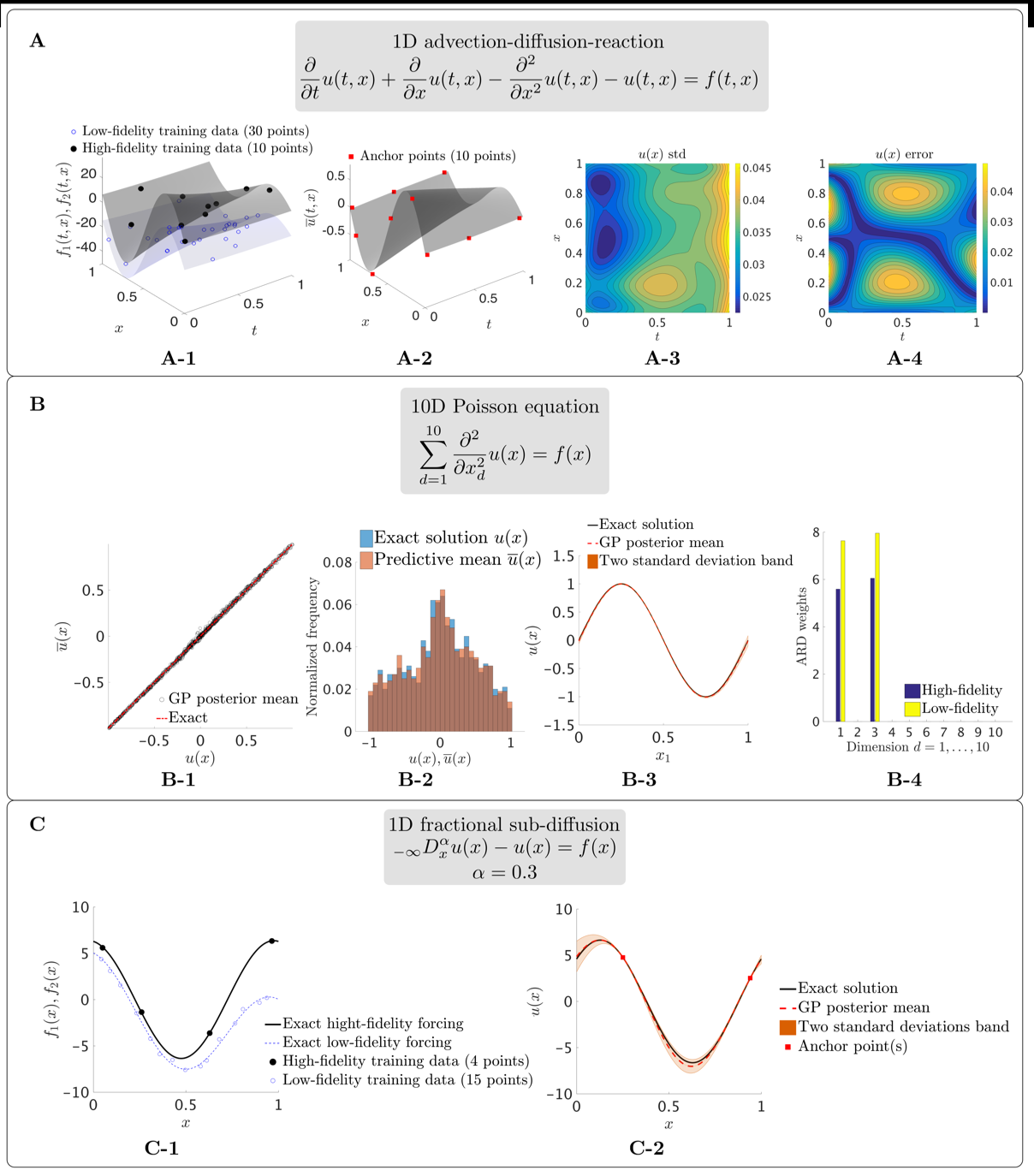}
\caption{{\em Generality and scalability of the multi-fidelity learning scheme}: Equations, variable fidelity data, and inferred solutions for a diverse collection of benchmark problems. In all cases, the algorithm provides an agnostic treatment of temporal integration, high-dimensionality, and non-local interactions, without requiring any modification of the workflow. Comparison between the inferred and exact solutions $\overline{u}$ and $u$, respectively, for (A) time-dependent advection-diffusion-reaction, (B) Poisson equation in ten dimensions, and (C) Fractional sub-diffusion.}
\label{fig:fig3}
\end{figure}

Our last example summarized in Figure \ref{fig:fig3}(C) involves a linear equation with fractional-order derivatives.  Such operators often arise in modeling anomalous transport, and their non-local nature poses serious computational challenges as it involves costly convolution operations for resolving the underlying non-Markovian dynamics \cite{podlubny1998fractional}. Bypassing the need for numerical discretization, our regression approach overcomes these computational bottlenecks, and can seamlessly handle all such linear cases without any modifications.
To illustrate this, consider the following one dimensional fractional equation

$$
{}_{-\infty}D^\alpha_x u(x) - u(x) = f(x),
$$
where $\alpha\in\mathbb{R}$ is the fractional order of the operator that is defined in the Riemann-Liouville sense \cite{podlubny1998fractional}. 
In our framework, the only technicality introduced by the fractional operators has to do with deriving the kernel $k(x,x';\theta)$. Here, $k(x,x';\theta)$ was obtained by taking the inverse Fourier transform \cite{podlubny1998fractional}
\[
[(-iw)^{\alpha}(-iw')^{\alpha} - (-iw)^{\alpha} - (-iw')^{\alpha} + 1]\hat{g}(w,w';\theta), 
\]
where $\hat{g}(w,w';\theta)$ is the Fourier transform of the kernel $g(x,x';\theta)$. Let us now assume that the low- and high-fidelity data $\{\mathbf{x}_1,\mathbf{y}_1\},\{\mathbf{x}_2,\mathbf{y}_2\}$ are generated according to $\mathbf{y}_\ell = f_\ell(\mathbf{x}_\ell) + \mathbf{\epsilon}_\ell$ where $\ell=1,2$, $\mathbf{\epsilon_1} \sim \mathcal{N}(0,0.3\ I)$, $\mathbf{\epsilon_2} \sim \mathcal{N}(0,0.05\ I)$, $f_2(x) = 2 \pi \cos(2 \pi x) - \sin(2 \pi x)$, and $f_1(x) = 0.8 f_2(x) - 5 x$. The training data $\mathbf{x}_1,\mathbf{x}_2$ with sample sizes $n_1 = 15, n_2 = 4$, respectively, are randomly chosen in the interval $[0,1]$ according to a uniform distribution. We also assume that we have access to data $\{\mathbf{x}_0,\mathbf{y}_0\}$ on $u(x)$. In this example, we choose $n_0 = 2$ random points in the interval $[0,1]$ to define $\mathbf{x}_0$ and let $\mathbf{y}_0 = u(\mathbf{x}_0)$. Notice that
\[
u(x) = \frac12 e^{-2 i \pi x} \left(  \frac{-i + 2 \pi}{-1 + (-2 i \pi)^\alpha} + \frac{e^{4 i \pi x} (i + 2 \pi)}{-1 + (2 i \pi)^\alpha}\right),
\]
is the exact solution, and is obtained using Fourier analysis. Our numerical demonstration corresponds to $\alpha = 0.3$, and our results are summarized in Figure 3(C).

\section{Discussion}
In summary, we have presented a probabilistic regression framework for learning solutions to general linear integro-differential equations from noisy data. Our machine learning approach can seamlessly handle spatio-temporal as well as high-dimensional problems. The proposed algorithms can learn from scattered noisy data of variable fidelity, and return solution fields with quantified uncertainty. This methodology generalizes well beyond the benchmark cases presented here. For example, it is straightforward to address problems with more than two levels of fidelity, variable coefficients, complex geometries, non-Gaussian and input-dependent noise models (e.g., student-t, heteroscedastic, etc. \cite{Rasmussen06gaussianprocesses}), as well as more general linear boundary conditions, e.g., Neumann, Robin, etc. The current methodology can be readily extended to address applications involving characterization of materials, tomography and electrophysiology, design of effective metamaterials, etc. An equally important direction involves solving systems of linear partial differential equations, which can be addressed using multi-output GP regression \cite{osborne2008towards,alvarez2009sparse}. Another key aspect of this Bayesian mindset is the choice of the prior. Here, for clarity, we chose to start from the most popular Gaussian process prior available, namely the stationary squared exponential covariance function. This choice limits our approximation capability to sufficiently smooth functions. However, one can leverage recent developments in {\em deep learning} to construct more general and expressive priors that are able to handle discontinuous and non-stationary response \cite{damianou2015deep,hinton2008using}. Despite its generality, the proposed framework does not constitute a universal remedy. For example, the most pressing open question is posed by non-linear operators for which assigning GP priors on the solution may not be a reasonable choice. Some specific non-linear equations can be transformed into systems of linear equations -- albeit in high-dimensions \cite{zwanzig1960ensemble,chorin2000optimal,denisov2009generalized} -- that can be solved with extensions of the current framework.

\section*{Acknowledgements}
We gratefully acknowledge support from DARPA grant N66001-15-2-4055. We would also like to thank Dr. Panos Stinis (PNNL) for the stimulating discussions during the early stages of this work.

\appendix

\section{Computer software} 
All data and results presented in the manuscript can be accessed and reproduced using the Matlab code provided at:\\
\url{https://www.dropbox.com/sh/zt488tymtmfu6ds/AADE2_Yb2Fz8AGBdsUmBXAyEa?dl=0}

\section{Movie S1}
We have generated an animation corresponding to the convergence properties of active learning procedure (see Figure 2). The movie contains 5 panels. The smaller top left panel shows the evolution of the computed posterior variance of $u$, while the smaller top right panel shows the corresponding error against the exact solution. Similarly, the smaller bottom left and bottom right panels contain the posterior variance and corresponding relative error in approximating the forcing term $f$. To highlight the chosen data acquisition criterion (maximum posterior variance of $f$) we have used a different color-map to distinguish the computed posterior variance of $f$. Lastly, the larger plot on the right panel shows the convergence of the relative error for both the solution and the forcing as the number of iterations and training points is increased. Figure 2 shows some snapshots of this animation.


%% References


%\bibliography{sample.bib}

\section*{References}

\begin{thebibliography}{10}

\bibitem{mumford2000dawning}
Mumford D (2000) The dawning of the age of stochasticity.
\newblock {\em Mathematics: frontiers and perspectives} pp. 197--218.

\bibitem{ghahramani2015probabilistic}
Ghahramani Z (2015) Probabilistic machine learning and artificial intelligence.
\newblock {\em Nature} 521(7553):452--459.

\bibitem{jordan2015machine}
Jordan M, Mitchell T (2015) Machine learning: Trends, perspectives, and
  prospects.
\newblock {\em Science} 349(6245):255--260.

\bibitem{diaconis1988bayesian}
Diaconis P (1988) Bayesian numerical analysis.
\newblock {\em Statistical decision theory and related topics IV} 1:163--175.

\bibitem{poincare1912calcul}
Poincar{\'e} H (1912) {\em Calcul des probabilit{\'e}s}.
\newblock (Gauthier-Villars).

\bibitem{hennig2015probabilistic}
Hennig P, Osborne MA, Girolami M (2015) Probabilistic numerics and uncertainty
  in computations in {\em Proc. R. Soc. A}.
\newblock (The Royal Society), Vol.{} 471, p. 20150142.

\bibitem{owhadi2015bayesian}
Owhadi H (2015) Bayesian numerical homogenization.
\newblock {\em Multiscale Modeling \& Simulation} 13(3):812--828.

\bibitem{sarkka2011linear}
S{\"a}rkk{\"a} S (2011) Linear operators and stochastic partial differential
  equations in {G}aussian process regression in {\em Artificial Neural Networks
  and Machine Learning--ICANN 2011}.
\newblock (Springer), pp. 151--158.

\bibitem{Rasmussen06gaussianprocesses}
Rasmussen CE (2006) Gaussian processes for machine learning.
\newblock (MIT Press).

\bibitem{murphy2012machine}
Murphy KP (2012) {\em Machine learning: a probabilistic perspective}.
\newblock (MIT press).

\bibitem{kennedy2000predicting}
Kennedy MC, O'Hagan A (2000) Predicting the output from a complex computer code
  when fast approximations are available.
\newblock {\em Biometrika} 87(1):1--13.

\bibitem{le2013multi}
Le~Gratiet L (2013) Ph.D. thesis (Universit{\'e} Paris-Diderot-Paris VII).

\bibitem{cohn1996active}
Cohn DA, Ghahramani Z, Jordan MI (1996) Active learning with statistical
  models.
\newblock {\em Journal of artificial intelligence research}.

\bibitem{krause2007nonmyopic}
Krause A, Guestrin C (2007) Nonmyopic active learning of {G}aussian processes: an
  exploration-exploitation approach in {\em Proceedings of the 24th
  international conference on Machine learning}.
\newblock (ACM), pp. 449--456.

\bibitem{mackay1992information}
MacKay DJ (1992) Information-based objective functions for active data
  selection.
\newblock {\em Neural computation} 4(4):590--604.

\bibitem{podlubny1998fractional}
Podlubny I (1998) {\em Fractional differential equations: an introduction to
  fractional derivatives, fractional differential equations, to methods of
  their solution and some of their applications}.
\newblock (Academic press) Vol.{} 198.

\bibitem{osborne2008towards}
Osborne MA, Roberts SJ, Rogers A, Ramchurn SD, Jennings NR (2008) Towards
  real-time information processing of sensor network data using computationally
  efficient multi-output {G}aussian processes in {\em Proceedings of the 7th
  international conference on Information processing in sensor networks}.
\newblock (IEEE Computer Society), pp. 109--120.

\bibitem{alvarez2009sparse}
Alvarez M, Lawrence ND (2009) Sparse convolved {G}aussian processes for
  multi-output regression in {\em Advances in neural information processing
  systems}.
\newblock pp. 57--64.

\bibitem{damianou2015deep}
Damianou A (2015) Ph.D. thesis (University of Sheffield).

\bibitem{hinton2008using}
Hinton GE, Salakhutdinov RR (2008) Using deep belief nets to learn covariance
  kernels for {G}aussian processes in {\em Advances in neural information
  processing systems}.
\newblock pp. 1249--1256.

\bibitem{zwanzig1960ensemble}
Zwanzig R (1960) Ensemble method in the theory of irreversibility.
\newblock {\em The Journal of Chemical Physics} 33(5):1338--1341.

\bibitem{chorin2000optimal}
Chorin AJ, Hald OH, Kupferman R (2000) Optimal prediction and the {M}ori--{Z}wanzig
  representation of irreversible processes.
\newblock {\em Proceedings of the National Academy of Sciences}
  97(7):2968--2973.

\bibitem{denisov2009generalized}
Denisov S, Horsthemke W, H{\"a}nggi P (2009) Generalized {F}okker-{P}lanck
  equation: Derivation and exact solutions.
\newblock {\em The European Physical Journal B} 68(4):567--575.

\bibitem{liu1989limited}
Liu DC, Nocedal J (1989) On the limited memory {BFGS} method for large scale
  optimization.
\newblock {\em Mathematical programming} 45(1-3):503--528.

\bibitem{snelson2005sparse}
Snelson E, Ghahramani Z (2005) Sparse {G}aussian processes using pseudo-inputs in
  {\em Advances in neural information processing systems}.
\newblock pp. 1257--1264.

\bibitem{hensman2013gaussian}
Hensman J, Fusi N, Lawrence ND (2013) Gaussian processes for big data.
\newblock {\em arXiv preprint arXiv:1309.6835}.


\end{thebibliography}



\end{document}